## Research

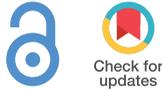

**Author for correspondence:**
Michael Veale
e-mail: m.veale@ucl.ac.uk

# Algorithms that remember: model inversion attacks and data protection law

## Michael Veale[1], Reuben Binns[2] and Lilian Edwards[3]

[1]Department of Science, Technology, Engineering and Public Policy, University College London, London, UK
[2]Department of Computer Science, University of Oxford, Oxford, UK
[3]Newcastle Law School, Newcastle University, Newcastle, UK

MV, 0000-0002-2342-8785

Many individuals are concerned about the governance of machine learning systems and the prevention of algorithmic harms. The EU's recent General Data Protection Regulation (GDPR) has been seen as a core tool for achieving better governance of this area. While the GDPR does apply to the use of models in some limited situations, most of its provisions relate to the governance of personal data, while models have traditionally been seen as intellectual property. We present recent work from the information security literature around 'model inversion' and 'membership inference' attacks, which indicates that the process of turning training data into machine-learned systems is not one way, and demonstrate how this could lead some models to be legally classified as personal data. Taking this as a probing experiment, we explore the different rights and obligations this would trigger and their utility, and posit future directions for algorithmic governance and regulation.

This article is part of the theme issue 'Governing artificial intelligence: ethical, legal and technical opportunities and challenges'.

## 1. Introduction

A recent heated topic in technology law and policy is whether machine learning systems are sufficiently regulated, a worry compounded by their apparent potential to reproduce societal discrimination and to transform mundane personal data into sensitive, invasive insight. The recent EU General Data Protection Regulation (GDPR), which strengthens data protection provisions and penalties, has been looked to internationally as a way forward, particularly following high-profile coverage of the scandal around Facebook and Cambridge Analytica in early 2018. Yet given that the trigger for data protection law is the processing of personal data rather than aggregated, 'anonymous' analytic products alone, the extent to which it is normally thought to provide control over models themselves is limited [1]. Instead, models have been typically thought to be primarily governed by varying intellectual property rights such as trade secrets, and are usually discussed as such [2].

In this paper, we argue that changing technologies may render the situation less clear-cut. We highlight a range of recent research which indicates that the process of turning training data into machine-learned systems is not one way but two: that training data, a semblance or subset of it, or information about who was in the training set, can in certain cases be reconstructed from a model. The consequences of this for the regulation of these systems is potentially hugely significant, as the rights and obligations for personal data differ strongly from the few generally thought applicable to models.

First, we introduce relevant elements of data protection law for a broad audience, contextualized by both current debates around algorithmic systems and the growing trend to trade and enable access to models, rather than to share underlying datasets. Second, we introduce model inversion and membership inference attacks, describing their set-up, and explain why data protection law is likely to classify vulnerable models as personal data. We then describe selected consequences of this change for data subjects, who would have access to new information, erasure and objection rights, and for the modellers and model recipients, who would need to consider security, 'data protection by design' and storage limitation implications. We conclude with a short discussion about the utility of this approach, and reflections on what this set-up tells us about desirable directions for future law-making around fast-moving technological practices.

## 2. European data protection law and machine learning

The GDPR is a lengthy and complex law, and is challenging to concisely summarize. It applies whenever personal data are processed (including collected, transformed, consulted or erased) either within the Union or outside the Union when the data relate to an EU resident. Personal data are defined by how much they can render somebody identifiable—going beyond email or phone number to include dynamic IP addresses, browser fingerprints or smart meter readings. The individual who the data relate to is called the *data subject*. The entities that determine 'the purposes and means of the processing of personal data' are the *data controllers*. Data subjects have rights over personal data, such as rights of access, erasure, objection to processing and portability of their data in a common machine-readable format elsewhere. Data controllers are subject to a range of obligations, such as ensuring confidentiality, notifying if data are breached and undertaking risk assessments. Additionally, data controllers must only process data where they have a lawful basis—such as consent—to do so, for a specified and limited purpose and for a limited period of storage.

Data protection law already governs the collection and use of data in generating machine learning models, and, under certain limited conditions, the application of model results upon data subjects. For example, (i) models cannot be trained from personal data without a specific lawful ground, such as consent, contract or legitimate interest; (ii) data subjects should be informed of the intention to train a model and (iii) usually maintain a right to object or withdraw consent; and (iv) in situations where models inform a significant, solely automated decision, individuals can appeal to the data controller for meaningful information about the logic of processing, or to have the decision taken manually reconsidered. The use of machine learning to turn 'normal' personal data into 'special category' personal data, such as race, political opinion or data concerning health, also requires establishing a lawful basis, which will usually be more stringent than personal data in general. The nature and utility of these provisions is explored and debated extensively elsewhere [3–6]; for reasons of space we do not expand upon them here.

## (a) Limits of the harms addressed by the current regime

If data protection already implicates machine learning in the above ways, does it matter if a model is personal data or not? In short—yes. The GDPR does touch upon machine learning, but largely indirectly, governing systems only when personal data are involved in training them or querying them to apply their results. Yet provisions would have a significantly different effect were models *themselves* to benefit from some of the status that personal data has. As it stands, there are no data protection rights nor obligations concerning models in the period *after* they have been built, but *before* any decisions have been taken about using them. Even the provisions that apply outside this period are relatively minor: those models being used as decision support, rather than as decision-making instruments, or those significantly affecting groups rather than individuals, are subject to few solid protections at all [4].

There is a reason to believe this level of control is insufficient. Individuals might want to control how they specifically are 'read' by machine-learned systems [6], particularly if individuals subscribe to the German belief in a right to informational self-determination (*informationelle Selbstbestimmung*). Trained models can transform seemingly non-sensitive data, such as gait or social media use, into sensitive data, such as information on an individual's fitness or medical conditions. In many fields, these models are far from user independent, and so individuals adding their own granular training data to these models enables them to predict more accurately in the future. Take automated lip-reading systems. These transform videos of speech into approximated transcripts, and, being largely speaker dependent [7], require individual-specific training data to be effective. Once a model integrates an individual's data, it will predict their speech with significantly greater accuracy than that of others. Do individuals have a right to reverse this process, and have some agency over the model after it has been trained and potentially traded? Do they even have some right to use the model, or at least to know what it is used for, given their stake in training it?

In a similar vein, arguments have been forwarded that 'groups' of individuals deserve agency over their representation in models. Scholars concerned with 'categorical' or 'group' privacy see such groups as having collective rights to determine how their identities are constituted [8,9]. Within this view, an ethical breach might be considered to have occurred 'when data or information is added to subject's identity without consent' [10]. For example, given claims of correlations between smartphone-captured data and aspects of physical/mental health [11], individuals revealing a rare condition to a modeller might have unintentionally enabled such inferences to be successfully made for others, too. Similar arguments also exist in relation to data that connect individuals more deeply, such as genomic data, where sequencing the genome of one family member might reveal information about many.[1]

## (b) Models on the move

These issues are of increasing importance given how data controllers increasingly refrain from trading data, as the ability to do this freely is heavily limited by data protection law, and instead look to trade or rent out models trained on it, as a way to pass on the value with fewer privacy and regulatory concerns. Many large firms already offer trained models for tasks including face recognition, emotion classification, nudity detection and offensive text identification. Two main business models underpin this practice.

The first is in the *licensing of application programming interfaces* (*APIs*) through 'App Store'-like platforms. Firms earn royalties when their models are deployed. Microsoft's *Cortana Intelligence Gallery* and Algorithmia's *Marketplace* match API providers with potential customers. Google's *Cloud AutoML* uses transfer learning to allow firms to enrich Google's generic models with their own specialized datasets. The trend towards augmentable, pre-trained 'learnware' [12] appears to be accelerating.

---

[1]A further related concern surrounds what one might call 'fair trade algorithms'—that using a system with data that have been collected unethically, perhaps in a country with limited fundamental rights or privacy rules, is itself unethical. As the users of such a system are unlikely to be in the training set, we do not address this issue here.

The second is the trading of *packaged models*. This might be preferable where APIs over the Internet are too sluggish, where queries are highly sensitive (e.g. medical records) or where transparency requirements require full access to model specifications. Apple's recent *Core ML* library and Google's *Tensorflow Mobile* are designed to run pre-trained models, some of which they provide, on portable devices.

Model trading is an attractive prospect in relation to many aspects of privacy. It stands in stark contrast to the incumbent (and, many suspect, largely illegal) system of large-scale, indefinite data accumulation and retention by shadowy and distant data brokers—systems presenting significant difficulties for data subject comprehension and control [13,14]. While it is commonly claimed that the 'free flow' of data is of economic benefit, much of this benefit derives from the movement of mined insights rather than the transmission of individual records. Given that the push towards open data might, in at least some regards, be at tension with privacy [15], model trading might serve as a useful approach to balance trade-offs that emerge.

Furthermore, model trading might mitigate concerns around platform monopolies. Enabling users to deploy local personalization tools might balance power relations against large firms hoarding personal data. This is the vision of *personal data container* proponents, who see distributed local processing, powered by decentralized, privacy-preserving analytical techniques—such as secure multi-party computation and homomorphic encryption—as enabling a shift of economic incentives and control from large players back to data subjects [16]. Many processes that use highly sensitive or granular knowledge have been envisaged as better managed using edge computing, such as the delivery of advertisements or the personalization of news media [17], or discrimination auditing and 'de-biasing' of machine learning models [18], limiting the sensitive data leaving devices users directly control.

Yet such systems, no longer representing unique records which might render an individual identifiable, have not been considered as personal data, and thus have been considered excluded from the data protection regime. In this paper, we challenge that conventional understanding, and reflect upon the legal provisions this would trigger. Recent evidence, reviewed here, highlights that models themselves may leak data they were trained with—raising classic data confidentiality concerns. Data protection rights and obligations might then apply to models themselves. We next outline the format model inversion attacks can take, why they might render models as personal data in the sense of European protection law and what the consequences of this might be for data subjects and for data controllers.

## 3. Why might models be personal data?

It has been demonstrated that machine learning models are vulnerable to a range of cybersecurity attacks that cause breaches of confidentiality. Confidentiality attacks leak information to entities other than those whom designers intended to view it. In the case of machine learning systems, there are different types of attacks. The first concerns *model stealing*, e.g. where an attacker uses API access to replicate a model [19]. Without a further confidentiality breach, this is primarily a concern for intellectual property rather than for privacy, and is of less concern here. A second attack class, *model inversion*, turns the journey from training data into a machine-learned model from a one-way one to a two-way one, permitting the training data to be estimated with varying degrees of accuracy. A third attack class, *membership inference*, does not recover the training data, but instead recovers information about whether or not a particular individual was in the training set. Both model inversion and membership inference can be undertaken as a *black-box attack*, where the attack can be done with only query access (e.g. through the API business model above), or as a *white-box attack*, where an attacker requires full access to the model's structure and parameters [20,21].

We will formally describe both model inversion and membership inference attacks in a manner amenable to the discussion of personal data. The set-up is as follows. A data controller holds a dataset of personal data $\mathbf{A} = a_{i,j} \in \mathbb{R}^{m \times n}$, where each $[a_{1,\star}, a_{2,\star} \dots a_{m,\star}]$ within is a row of personal characteristics relating to one of the $m$ data subjects in the set $|DS_1| = m$, with each of the $n$ variables indexed by $j$. They also have access to a model $M(\mathbf{B})$, which is a machine-learned model trained on personal data $\mathbf{B} = b_{i,j} \in \mathbb{R}^{x \times y}$, where each $[b_{1,\star}, b_{2,\star} \dots b_{x,\star}]$ within is a row of personal characteristics relating to one of $x$ data subjects in the set $|DS_2| = x$, and each one of the $y$ variables is a feature in the trained model. The data controller may have access to the model either directly (white box) or via a query interface (black box). We assume $DS_1 \cap DS_2 > 0$: that is, some individuals are in both the training set and the additional dataset held. We refer to individuals in both $\mathbf{A}$ and $\mathbf{B}$ as being in set $Z$.

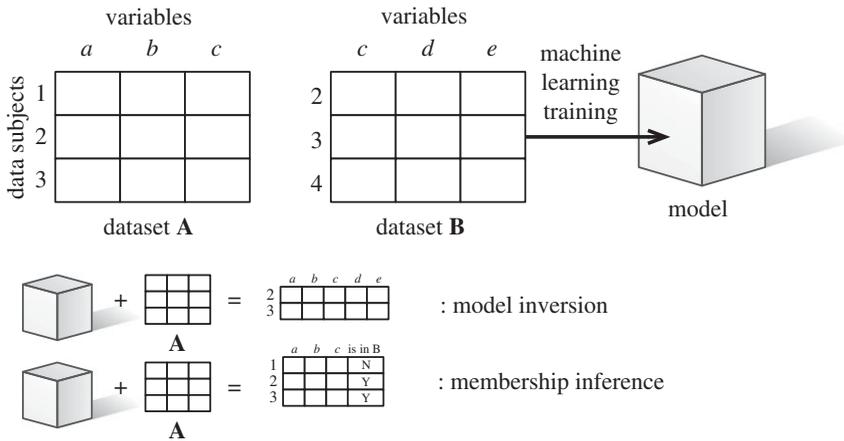

**Figure 1.** Model inversion and membership inference attacks.

## (a) Attack types

In the following section, we outline, at a high level and non-exhaustively, forms and work on model inversion and membership inference attacks. A visual depiction of both can be found in figure 1.

### (i) Model inversion

Under a model inversion attack, a data controller who does not initially have direct access to $\mathbf{B}$ but is given access to $\mathbf{A}$ and $M(\mathbf{B})$ is able to recover some of the variables in training set $\mathbf{B}$, for those individuals in both the training set and the extra dataset $\mathbf{A}$. These variables connect to each other, such that the new personal dataset in question has all the variables of $\mathbf{A}$ and some of $\mathbf{B}$. There may be error and inexactitude in the latter, but the data recovered from those in the training dataset will be more accurate than characteristics simply inferred from those that were not in the training dataset.

One of the earliest attacks resembling model inversion emerged from the field of recommender systems; a demonstration that collaborative filtering systems, where item recommendations are generated for a user based on behavioural patterns of other users, can end up revealing the consumption patterns of individual users [22]. At the time, a reader might have reasonably assumed such risks to be a quirk of the particular system and application area, but subsequent work suggests that this may be a more general problem potentially facing any kind of machine learning model that uses personal data for training. Later work undertook attacks against several machine learning models to attempt to learn meaningful information about the training data. They were not concerned with privacy *per se* but with trade secrets, seeking to uncover the 'secret sauce' of algorithms that might give them a commercial edge, such as whether speech recognition systems were trained on certain accents or not. Ateniese *et al.* [23] and Fredrikson *et al.* [24] examined models designed to select correct medical doses for a widely used anticoagulant that interacts strongly with individual genetic markers. These authors demonstrated the possibility of reverse engineering to reveal patients' genetic markers with some demographic information about patients in the training data.[2] Further work demonstrated both white- and black-box attacks re-identifying individuals from models trained on survey data with no false positives, and black-box attacks to reconstruct faces from facial recognition systems to the point where skilled crowdworkers could use the photo to identify an individual from a line-up with 95% accuracy [20].

Other work connected to model inversion has indicated that only small changes to training algorithms lead to nearly indistinguishable models that it is possible to exploit to leak large amounts of private data [26], or that systems exactly memorize specific private information in training sets, such as strings of sensitive data [27]. Some model structures also require some training data in order to predict. An example is the support vector machine family, where the training vectors themselves that 'support' the classification boundary are bundled with the model used.[3]

### (ii) Membership inference

Membership inference attacks do not recover training data, but instead ascertain whether a given individual's data were in a training set or not. Under a membership inference attack, the holder of **A** and M(**A**) does not recover any of the columns in **B**, but can add an additional column to dataset **A** representing whether or not a member of $DS_1$ is in the set $Z$: that is, whether or not they were also part of the training set participants $DS_2$.

Shokri *et al.* [28] demonstrated membership inference in a black-box attack against a hospital discharge model. Connectedly, Pyrgelis *et al.* [29] looked at membership inferences in location data, showing that it is comparatively easy to examine whether an individual with a certain movement pattern was used in the construction of an aggregate. They note this could be concerning if such aggregates were themselves made to understand a sensitive subgroup, such as individuals with dementia.

## (b) Why do these attacks render models as personal data?

While these attacks are nascent, their potential is being increasingly demonstrated.[4] The initial important question for data protection law is: to what extent would systems vulnerable to attacks like these be considered datasets of personal data?

There are strong arguments that a model inversion–vulnerable M(**B**) would be personal data. A direct analogy can be made to personal data which have been 'pseudonymized'. Article 4(5) of the GDPR defines pseudonymization as 'the processing of personal data in such a manner that the personal data can no longer be attributed to a specific data subject without the use of additional information, provided that such additional information is kept separately and is subject to technical and organizational measures to ensure that the personal data are not attributed to an identified or identifiable natural person'. Under European law, pseudonymized data explicitly remain personal data. In the above set-up, M(**B**) represents the pseudonymized version of the training set **B**, while **A** represents the key by which such data can be partially re-identified. Where a single data controller is in possession of both **A** and M(**B**), and a model inversion attack is possible, it would appear by analogy that not only **A** but also M(**B**) should be considered personal data.

---

[2]Some argue that this paper shows no privacy breach here beyond people finding machine learning 'creepy'; see [25].

[3]Thanks to Josh Kroll for bringing this point to our attention.

[4]We are, however, cautious about overstating their practical efficacy, which does remain unclear. We are not aware of any documented attacks 'in the wild'.

Of potentially greater interest, however, is the situation where a model M(**B**) has been released, and so **A** and M(**B**) are held by different entities. There is legal precedent for the model M(**B**) to be considered personal data in this case too.

'Personal data' means any information relating to an identified or identifiable natural person (Article 4(1)). The surprising ease with which individuals can be identified in supposedly anonymous datasets [30] creates a great deal of uncertainty around which datasets are not personal data [31]. Recital 26 provides an aide to navigating this problem, with a test of reasonable likelihood of re-identification.

> To determine whether a natural person is identifiable, account should be taken of all the means reasonably likely to be used, such as singling out, either by the controller or by another person to identify the natural person directly or indirectly. To ascertain whether means are reasonably likely to be used to identify the natural person, account should be taken of all objective factors, such as the costs of and the amount of time required for identification, taking into consideration the available technology at the time of the processing and technological developments.

When asked in recent years, the Court of Justice of the European Union (CJEU) has set a low bar and wide scope for what is considered personal data. In *Breyer* (ECLI:EU:C:2016:779), the CJEU clarified the reasonable likelihood test in the Data Protection Directive 1995 that now forms part of Recital 26 of the GDPR. In particular, the Court took a wide view of identifiability, clarifying that where additional data are required to re-identify a dataset, those data need not all be in the hands of a single entity to be considered personal. In our case, this illustrates that in the case of either attack M(**B**) might be considered personal data if dataset **A** is held by another entity.[5]

Furthermore, while the data returned from model inversion attacks are quite easily construed as personal data, insofar as they resemble a training set or can be used to identify individuals, it might initially appear less clear that data obtained from membership inference—whether an individual was in a training dataset—could be. 'Personal data' does not specify particular sensitive or private categories in the definition, instead using the phrase 'any information'. In the recent CJEU case *Nowak* (ECLI:EU:C:2017:994), the Court engaged with the meaning of this phrase, noting that it is 'not restricted to information that is sensitive or private, but potentially encompasses all kinds of information, not only objective but also subjective'. Similarly, and following the Article 29 Working Party, the pan-European group of regulators that worked together to provide guidance on the Data Protection Directive, the Court understood that this 'any information' can 'relate to' an individual in many ways: by content, by purpose and by effect.[6] Seen through this lens, information on an individual's membership of a training set would indeed fall within the scope of personal data, regardless of how trivial or mundane it might be to the individual it concerns.

There exists an argument forwarded by scholars such as Purtova [31] which claims that the Court's approach to the scope of personal data fuels undesirable data protection maximalism. She considers a situation where common environmental factors which do not identify an individual by themselves, such as the weather, are used in a smart city context as an input to a behavioural profiling system. In this case, she argues, it is plausible that the weather data would be personal data by means of purpose and effect. In the context of this paper, her argument might plausibly be extended to claim that the weights, and perhaps the structure, of machine learning models relate to an individual by means of impact, and by virtue of this are personal data. While we acknowledge this *reductio ad absurdum* argument concerning the current scope of personal data, and the consequences for it as making the law impractically broad, our argument does not lean in this direction. We do not aim to support, oppose or resolve the dilemma raised by Purtova;

---

[5]For a contrasting view of anonymization, which considers data as 'functionally anonymous' where we go beyond the data themselves to also consider the surrounding environment, see [32].

[6]This specific opinion document has not been explicitly carried forward by the European Data Protection Board, which inherits the mantle of the Article 29 Working Party at the time of writing; however, the Court in *Nowak* also noted that information can relate to an individual by 'content, purpose of effect' (Paragraph 44).

but merely to note that the argument made here—that inverted models might fall under the definition of personal data—does *not* depend on the kind of expansive definition that gives rise to such absurdities. Thus, even if the definition were to be somehow tightened in scope (indeed, the scope of personal data has changed even between recent cases such as *YS and others* (ECLI:EU:C:2014:2081) and *Breyer*), the argument above concerning inverted models would still probably stand.

In sum, model inversion and membership inference attacks, where possible, do risk models being considered as personal data even without resorting to a maximalist reading of data protection law. The question then remains—what are the practical consequences of this, and are they of interest to those hoping to better control these systems and hold them to account?

## 4. Implications of models as personal data

Were models to be personal data, very different sets of rights and obligations would be triggered compared with their status today. In this section, we survey a salient range of them from two angles—rights that would be newly afforded to the data subject, and obligations placed on the data controller.

### (a) For data subjects

In the light of these attacks, we now consider the viability of three tasks data subjects might want to achieve in relation to models trained with their data: where a data subject wishes to access models and to know where they have originated from and to whom they are being traded or transmitted; where a data subject wishes to erase herself from a trained model; and where a data subject wishes such a model not to be used in the future.

### (i) Information rights

The GDPR contains information provisions which are triggered on request, such as the right of access (Article 15) and the right of portability (Article 20), as well as provisions where the data controller must provide information to the data subject without being solicited to do so (Articles 13 and 14). In practice, these rights are most well known for the requirement, carried through from the Data Protection Directive, to provide all the information held about an individual upon request. It seems unlikely that such a request would even in theory allow an individual to request an entire model trained using multiple individuals on the basis that it contained that individual's data as to do so might compromise the privacy of others in the training set.[7] To provide it might even breach data protection's security principle. It also seems unlikely that the individual would easily succeed at requesting a copy of the data retrieved from a model, as the data are unlikely to include a name or identifier, and to make names and identifiers difficult to access is an increasingly common practice by data controllers, leaving personal data difficult to exercise rights over [33].

Particularly given the increasing interest in model trading described above, it is more interesting to consider other aspects of these information rights. These rights might enable better tracking of the provenance of data, analysis and decision-making systems, something which has received increasing attention of late [34]. In effect, they help track the origin and destination of trained models. The GDPR does have two core types of requirement which, respectively, require controllers with data to provide information about where specifically personal data came from, and to whom, more generally, the data are going.

At the point of collection (Article 13(1)(e)), and upon request (Article 15(1)(c)), the data subject should be provided with 'the recipients or categories of recipients of the personal data, if any'.

---

[7]A case where only one individual's data were used might include, for example, a tailored voice recognition system or a learned smart home personalization model. Portability of these models to another controller might be useful, although it would introduce a probably infeasible compatibility demand on the receiving provider, who would then probably find it easier simply to take the raw data and retrain a model themselves.

While data protection authorities' interpretations have indicated that providing potentially broad and unhelpful 'categories' alone is in line with the law [35], recent guidance from the Article 29 Working Party notes that [36] only providing the 'categories' will require a justification of why this is in line with the principle of fairness in data protection, and if categories alone are provided, they must 'be as specific as possible by indicating the type of recipient (i.e. by reference to the activities it carries out), the industry, sector and sub-sector and the location of the recipients.'

Where one controller receives personal data from another—as is the case with model trading if we conceive of models that way—there is an alternative approach afforded by Article 14(2)(f). This provision states that 'the controller shall provide the data subject with the following information necessary to ensure fair and transparent processing in respect of the data subject [· · · ] from which source the personal data originate, and if applicable, whether it came from publicly accessible sources'.

The only applicable exemption here is Article 14(5)(b): 'the provision of such information proves impossible or would involve a disproportionate effort'. It is highly likely that a controller faced with the situation of a model-as-personal-data would try to use this exemption, particularly as they are unlikely to have contact information for the data subjects in the model, and Article 11(1) notes they are not obliged to hold such additional data for the sole purposes of complying with the GDPR. This is, however, not the end of the story, because Article 14(5)(b) goes on to note that, where disproportionate effort applies, '[in] such cases the controller shall take appropriate measures to protect the data subject's rights and freedoms and legitimate interests, including making the information publicly available.' The one safeguard that is explicitly listed would seem to imply that the data controller might then be obliged to make publicly available— perhaps through its website—its receipt of a model and the sources from which the personal data originate.

In effect, this would mean that organizations receiving invertible or potentially invertible models would have to publish where they came from. This information would be a huge asset to those mapping these flows and looking to understand the data economy and algorithmic accountability within it, particularly if such notices were machine readable in form.

## (ii) Erasure rights

The famous 'right to be forgotten' is a qualified right of a data subject to 'the erasure of personal data concerning him or her' (Article 17). Core reasons a data subject might want to erase herself from a model overlap with the general reasons for model control presented earlier—to erase insights about her she might dislike; to erase unwanted insights about a group she identifies as being part of; or to erase insights which might lead to data breaches.

There are two main ways to erase data from a trained model. First, a model can be trained based upon an amended training dataset. The computational intensity of much machine learning training, even on the world's most powerful computational infrastructures, means such training is far from cost free or instantaneous. Retraining brings significant energy costs—data centres consume between 1.1% and 1.5% of global power consumption [37]—as well as time and labour costs.

Second, the model itself can be amended after training. This is not easy, and rarely currently possible in modern systems. Approaches for quick and easy 'machine unlearning' are only beginning to be proposed and are still largely unexplored, let alone at a stage ready for deployment [38,39]. Methods currently on the table cannot be retrofitted onto existing systems, and would require entire model pipelines to be re-conceived, with unclear effects. Consequently, large-scale organizations such as Google do not appear to remove links from their trained search model, but filter (delist) results between model output and search result delivery [40].

In many cases, an individual choosing to remove herself from the training set of a model would have little impact on the patterns the model has learned. Models basing patterns on single data records are generally considered to have been undesirably *overfitted*—memorizing the training data rather than finding generalizable patterns.[8] Such a right becomes more useful when used collectively. Whether this is possible will depend on both the phenomena at hand and the level of co-ordination between the data subjects. Co-ordination in the form of online petitions and crowdfunding make this an interesting avenue to explore for the right of erasure, and one that is already being probed in relation to the right of access [41].

### (iii) Right to restrict processing and to object

Under data protection law, individuals can also say no to types of processing which they did not consent to, such as those relying on 'legitimate interest' or a 'public task'. The right to object in Article 21 allows an individual to object to processing based on the legitimate interests of the data controller of public interest (largely public sector processing), as long as the data controller cannot demonstrate a 'compelling' legitimate interest of its own—something which seems a high bar indeed, and seemingly unlikely to be met unless some social benefit is present, rather than just controllers' economic incentive.

The right to restrict processing in Article 18 has a wider range of provisions. For our purposes, it gives the data subject rights to stop processing of their data until a balancing test relating to the right to object can be carried out, or until the accuracy of the data is ascertained and rectified. This is immediate but time limited, and so while it could in theory be used quite disruptively, it is generally considered the difficulties seeing models as personal data would engender.

If models were personal data, the consequences of these rights become confusing. What is it to object to the use of a model? To query a model might be considered analogous to consulting the personal dataset within. Consultation is one of the many actions that explicitly constitute processing in data protection. Yet because the data are not organized by record when they are in the form of the model, querying a model seems more like querying the entire dataset than querying a single record. Does objection give every individual a veto right over the consultation of the model in its entirety, for any lawful purpose? This would seem highly problematic, and probably disproportionate, but it is a possible reading of this provision, and one which demonstrates the difficulties seeing models as personal data would engender.

A more sensible approach from a user's perspective to achieve their goals would often be to restrict the processing of that model in relation to an individual decision, but this is already possible by using these rights to object or restrict the personal data constituting the query being used for prediction in a specific case, or possibly the right not to be subject to solely automated decision-making (Article 22), which, unusually for data protection, instead targets the decisions or measures themselves rather than the use of data. Yet the strange interplay between objection and models made of mixed-up personal data could potentially be a place of tension were individuals to try to test and enforce these rights.

## (b) For data controllers

Data controllers must consider both the specific rights and obligations they are subject to, as well as adhere to the overarching principles of data protection. We highlight two relevant areas here in the context of models as personal data: the security principle and the storage limitation principle.

### (i) Security principle and data protection by design

As an overarching concern and obligation, data controllers need to consider whether their systems leak personal data in the first place. It seems unlikely that a modeller would wish to establish a legal basis for onward transfer of a model they have trained, something which would be especially onerous if that model is being transmitted outside of the EU or a country deemed

---

[8]There are times when one record can make a difference: 'one-shot learning' in image recognition, for example. As children, humans are very good at seeing an object once, classifying it, and then correctly identifying a second, different object of the same classification. Standard machine learning systems perform very poorly at this task, requiring many examples of a single class in order to become good at recognizing it.

'adequate' with EU data protection law. If they do transfer such a model without a legal basis to do so, and such a model is inverted, it would probably be considered both a data breach and a violation of the security principle more generally. This is a relatively shocking conclusion, and endangers many of the positive aspects of model trading, particularly that large datasets are not being transmitted constantly, which this paradigm promises.

Further reason to ensure that models are not invertible comes from Article 25(1) of the GDPR, which introduces a qualified obligation to implement technical and organizational measures designed to implement data protection principles. In combination, these aspects require us to consider how models can be made which are resilient to the attacks described above.

Thankfully, security researchers try to secure systems as well as try to break them. The most common defence that can be levied against model inversion attacks is differential privacy [42]: a defence discussed in most of the papers above describing the threats. Differential privacy is a formalized notion of privacy as information disclosure.[9] At the core of differential privacy is the notion of adding precise forms of random noise to queries such that for every individual in a dataset, removing them from that dataset will not noticeably change the results of that query.[10] Differential privacy guarantees essentially limit what can be learned about any individual in the dataset to that which could be learned about them from everybody else's data alone. Differentially private approaches to machine learning work in a similar way: models lacking data relating to a particular individual function extremely similarly to those containing it [45]. This provides intrinsic protection against a great deal of model inversion attacks.

*Theoretically*, such learning algorithms are just as powerful as non-differentially private ones [46]. Since algorithms are supposed to find generalizable patterns, not memorize and regurgitate specific records, this makes intuitive sense (discussed further below). Yet despite the growing interest in differential privacy [47–49], the real challenge comes with deployment. The tools available today can be computationally expensive to deploy [50] as well easily undermined with even small or arcane software errors [51]. Only a few large and powerful companies have demonstrated an ability to deploy them, and only then for very limited purposes. Furthermore, differential privacy works well at protecting disclosure in contexts where every individual has the same weight, such as in a count, but poorly in situations where there are extreme outliers; for example, in a dataset of wealth containing Bill Gates, as the amount of noise needed to be added reduces the query results to absurdity [52,53]. In cases where outliers might be vulnerable as well as powerful this is problematic: we might suspect it is they who need data protection the most [4].

A second linked line of defence, albeit without the guarantees differential privacy provides, is to attempt to make models that do not 'overfit' to the data. Where they do, they are confusing the 'noise' in the dataset for the generalizable 'signal' that helps prediction on unseen cases: memorizing rather than doing anything close to learning. Avoiding overfitting is important—and, in this way, data protection by design might legally oblige the training of methodologically sound models—but avoiding it is not enough to guarantee a model will not be vulnerable to model inversion. In some cases, such attacks have been shown to succeed in part even in the absence of overfitting [54].

## (ii) Storage limitation principle

Relatedly, another data protection principle, that of 'storage limitation', applies in this case. Storage limitation means that data should be kept for no longer than is necessary for the purposes for which they are processed. As training data may need to be discarded as time goes on to meet this obligation, so might models. Some techniques from the field of concept drift adaptation could be useful here. This is the domain of research which looks at how to understand and model changing phenomena in machine learning systems. For example, Gama *et al.* [55] describe a variety of methods used to limit the use of older data in machine learning systems by both managing data used in machine learning and gradually forgetting data within a model. These methods are primarily used to better model phenomena today, particularly where old correlations may now be irrelevant; however, similar efforts are likely to be required for any systems for which model inversion is not a readily remediable vulnerability.

---

[9]Differential privacy was popularized by Dwork *et al.* [43]. A useful lay introduction is provided in [44].

[10]Pure differential privacy would mean that the analysis is *exactly* the same; in practice there is a small margin of flexibility that is allowed.

## 5. Discussion and conclusion

In this paper, we have outlined how recent confidentiality attacks upon machine learning systems, including model inversion and membership inference, interplay with data protection law. Where models are vulnerable to such attacks, they gain an additional dimension—not only an analytic product potentially protected by intellectual property rights, but also a set of personal data, conceptually close to the idea of 'pseudonymization' in the GDPR. We illustrated a selection of consequences that could flow from this new classification that are relevant to discussions of the governance of machine-learned models which are being traded and transferred between data controllers. These include those of direct utility to data subjects, such as information, erasure and objection rights, and overarching obligations relevant to data controllers, such as security and storage limitation provisions. Seeing models as personal data could serve to re-balance or at least disrupt the power relations between those holding models and those whose data are used to train them.

There is, however, reason for caution amidst the promise. While potentially enabling useful provisions, requiring what is essentially a security vulnerability in order to trigger rights and obligations is disconnected and arbitrary. Models not amenable to model inversion might still be models that individuals wish to know the origin or destination of or wish to have themselves or their group erased from. We suspect many situations of problematic profiling will not be vulnerable to model inversion, and therefore are not governable with this approach. More research on which types of model, including those trained on particular types of data, are vulnerable in practice and which are not is needed to help us understand the potential implications of these attacks. Furthermore, where some model inversion is possible, technical challenges will make it difficult for data subjects and regulators to prove the leakiness. In many cases, while a companion dataset that may enable such an attack might exist, potentially on shady markets, it will not be held by the data subject, auditor or the regulator, and thus the risk will be difficult to assess even if full model access is provided. The set-up appears to further burden every stakeholder apart from the modeller, and it remains questionable whether downstream governance provisions in general, such as rights after model training, are the best way to deal with algorithmic harms [4].

Data protection provisions were not designed with the profiling and inference capabilities we see today. It is becoming clearer that many socio-technical challenges presented by machine learning and algorithmic systems more broadly are not wholly dealt with using the provisions in regulations such as the GDPR, which are the result of a slow evolution in definitions and concerns. On one hand, this has created a set of principles that are desirable and robust for many purposes. On the other hand, it is showing that seeing data protection as an omnibus governing regime for all data-driven issues is misguided. In this paper, we illustrate this by considering model inversion as a regulatory experiment—one that is potentially realistic, although far from advisable as a foundation for future algorithmic governance. We argue it provides a hook to probe whether or not it is appropriate that these rights and obligations extend to analytic products themselves, and what the consequences of this development might be. Some consequences, such as the mapping of the provenance of trained models, seem potentially useful in oversight of increasingly influential systems. Others, such as the right to object and restrict processing, seem at tension with the very notion of model inversion. As the domain of personal data expands, it is important to recognize that, while its scope is wide, its scope of effective and enforceable governance might not be. Pushing the boundaries of data protection law in the light of new technologies serves as a useful and clarifying force in the continuous task of the coming decades in better applying existing law, and developing regulatory regimes around societal issues they fail to deal with in practice.

Data accessibility. This article has no additional data.

Competing interests. We declare we have no competing interests.

Funding. Supported by the Engineering and Physical Sciences Research Council (EPSRC): EP/M507970/1 (M.V.), EP/J017728/2 (R.B.), EP/G065802/1 (L.E.); and the Arts and Humanities Research Council (AHRC) AH/K000179/1 (L.E.).

# References

1. Hildebrandt M. 2008 Profiling and the rule of law. *Identity Inf. Soc.* **1**, 55–70. (doi:10.1007/s12394-008-0003-1)
2. Depreeuw S, de Vries K. 2016 *Deliverable 3.11: profile transparency, trade secrets and intellectual property rights in OSNs*. Brussels, Belgium: USEMP Project.
3. Wachter S, Mittelstadt B, Floridi L. 2017 Why a right to explanation of automated decision-making does not exist in the General Data Protection Regulation. *Int. Data Privacy Law* **7**, 76–99. (doi:10.1093/idpl/ipx005)
4. Edwards L, Veale M. 2017 Slave to the algorithm? Why a 'right to an explanation' is probably not the remedy you are looking for. *Duke Law Technol. Rev.* **16**, 18–84. (doi:10.31228/osf.io/97upg)
5. Selbst AD, Powles J. 2017 Meaningful information and the right to explanation. *Int. Data Privacy Law* **7**, 233–242. (doi:10.1093/idpl/ipx022)
6. Hildebrandt M. 2015 *Smart technologies and the end(s) of law*. Cheltenham, UK: Edward Elgar.
7. Bear HL, Cox SJ, Harvey RW. 2015 Speaker-independent machine lip-reading with speaker-dependent viseme classifiers. In *The 1st Joint Conf. on Facial Analysis, Animation, and Auditory-Visual Speech Processing* (*FAAVSP-2015*), *Vienna, Austria, 11–13 September 2015*, pp. 190–195. See https://www.isca-speech.org/archive/avsp15.
8. Vedder A. 1999 KDD: the challenge to individualism. *Ethics Inf. Technol.* **1**, 275–281. (doi:10.1023/A:1010016102284)
9. Taylor L, Floridi L, van der Sloot B. 2017 *Group privacy*. Berlin, Germany: Springer.
10. Mittelstadt B. 2017 From individual to group privacy in big data analytics. *Phil. Technol.* **30**, 475–494. (doi:10.1007/s13347-017-0253-7)
11. Farhan AA *et al.* 2016 Behavior vs. introspection: refining prediction of clinical depression via smartphone sensing data. In *Proc. of the 2016 IEEE Wireless Health* (*WH*), *Bethesda, MD, 25–27 October 2016*. New York, NY: IEEE.
12. Zhou ZH. 2016 Learnware: on the future of machine learning. *Front. Comput. Sci.* **10**, 589–590. (doi:10.1007/s11704-016-6906-3)
13. Van Kleek M, Liccardi I, Binns R, Zhao J, Weitzner DJ, Shadbolt N. 2017 Better the devil you know: exposing the data sharing practices of smartphone apps. In *Proc. of the 2017 CHI Conf. on Human Factors in Computing Systems* (*CHI'17*), *Denver, CO, 6–11 May 2017*, pp. 5208–5220. New York, NY: ACM.
14. Binns R, Lyngs U, Van Kleek M, Zhao J, Libert T, Shadbolt N. 2018 Third party tracking in the mobile ecosystem. In *Proc. of the 10th ACM Conf. on Web Science, Amsterdam, The Netherlands, 27–30 May 2018*. New York, NY: ACM.
15. Banisar D. 2011 *The right to information and privacy: balancing rights and managing conflicts*. Washington, DC: World Bank.
16. Crabtree A, Lodge T, Colley J, Greenhalgh C, Mortier R, Haddadi H. 2016 Enabling the new economic actor: data protection, the digital economy, and the databox. *Pers. Ubiquitous Comput.* **20**, 947–957. (doi:10.1007/s00779-016-0939-3)
17. Crowcroft J, Madhavapeddy A, Schwarzkopf M, Hong T, Mortier R. 2011 Unclouded vision. In *Distributed computing and networking* (eds MK Aguilera, H Yu, NH Vaidya, V Srinivasan, RR Choudhury), pp. 29–40. Berlin, Germany: Springer.
18. Kilbertus N, Gascon A, Kusner M, Veale M, Gummadi K, Weller A. 2018 Blind justice: fairness with encrypted sensitive attributes. In *Proc. of the 35th Int. Conf. on Machine Learning* (*ICML 2018*), *Stockholm, Sweden, 10–15 July 2018*. International Machine Learning Society.

19. Tramèr F, Zhang F, Juels A, Reiter MK, Ristenpart T. 2016 Stealing machine learning models via prediction APIs. In *USENIX Security Symp., Austin, TX, 10–12 August 2016*, pp. 601–618. See https://www.usenix.org/system/files/conference/usenixsecurity16/sec16_paper_tramer.pdf.

20. Fredrikson M, Jha S, Ristenpart T. 2015 Model inversion attacks that exploit confidence information and basic countermeasures. In *Proc. of the 22nd ACM SIGSAC Conf. on Computer and Communications Security, Denver, CO, 12–16 October 2015*, pp. 1322–1333. New York, NY: ACM.

21. Abadi M, Erlingsson U, Goodfellow I, McMahan HB, Mironov I, Papernot N, Talwar K, Zhang L. 2017 On the protection of private information in machine learning systems: two recent approaches. In *Proc. of the 30th IEEE Computer Security Foundations Symp., Santa Barbara, CA, 21–25 August 2017*. New York, NY: IEEE.

22. Calandrino JA, Kilzer A, Narayanan A, Felten EW, Shmatikov V. 2011 'You might also like:' privacy risks of collaborative filtering. In *Proc. of the 2011 IEEE Symp. on Security and Privacy (SP), Oakland, CA, 22–25 May 2011*, pp. 231–246. New York, NY: IEEE.

23. Ateniese G, Mancini LV, Spognardi A, Villani A, Vitali D, Felici G. 2015 Hacking smart machines with smarter ones: how to extract meaningful data from machine learning classifiers. *Int. J. Secur. Networks* **10**, 137–150. (doi:10.1504/IJSN.2015.071829)

24. Fredrikson M, Lantz E, Jha S, Lin S, Page D, Ristenpart T. 2014 Privacy in pharmacogenetics: an end-to-end case study of personalized warfarin dosing. *Proc. USENIX Secur. Symp.* **2014**, 17–32.

25. McSherry F. 2016 Statistical inference considered harmful. See https://github.com/frankmcsherry/blog/blob/master/posts/2016-06-14.md.

26. Song C, Ristenpart T, Shmatikov V. 2017 Machine learning models that remember too much. In *Proc. of the 2017 ACM SIGSAC Conf. on Computer and Communications Security (CCS '17), Dallas, TX, 30 October–3 November 2017*, pp. 587–601. New York, NY: ACM.

27. Carlini N, Liu C, Kos J, Erlingsson Ú, Song D. 2018 The secret sharer: measuring unintended neural network memorization & extracting secrets. (http://arxiv.org/abs/1802.08232)

28. Shokri R, Stronati M, Song C, Shmatikov V. 2017 Membership inference attacks against machine learning models. In *Proc. of the 2017 IEEE Symp. on Security and Privacy (SP), San Jose, CA, 22–24 May 2017*, pp. 3–18. New York, NY: IEEE.

29. Pyrgelis A, Troncoso C, De Cristofaro E. 2018 Knock knock, who's there? Membership inference on aggregate location data. In *Proc. of the Network and Distributed Systems Security (NDSS) Symposium 2018, San Diego, CA, 18–21 February 2018*. Reston, VA: The Internet Society.

30. Ohm P. 2009 Broken promises of privacy: responding to the surprising failure of anonymization. *UCLA Law Rev.* **57**, 1701–1777.

31. Purtova N. 2018 The law of everything. Broad concept of personal data and future of EU data protection law. *Law Innov. Technol.* **10**, 40–81. (doi:10.1080/17579961.2018.1452176)

32. Elliot M, O'Hara K, Raab C, O'Keefe CM, Mackey E, Dibben C, Gowans H, Purdam K, McCullagh K. 2018 Functional anonymisation: personal data and the data environment. *Comput. Law Secur. Rev.* **34**, 204–221. (doi:10.1016/j.clsr.2018.02.001)

33. Veale M, Binns R, Ausloos J. 2018 When data protection by design and data subject rights clash. *Int. Data Privacy Law* **8**, 105–123. (doi:10.1093/idpl/ipy002)

34. Singh J, Cobbe J, Norval C. 2018 Decision provenance: capturing data flow for accountable systems. (http://arxiv.org/abs/1804.05741)

35. Galetta A, de Hert P. 2017 Exercising access rights in Belgium. In *The unaccountable state of surveillance: exercising access rights in Europe* (eds C Norris, P de Hert, X L'Hoiry, A Galetta), pp. 77–108. Cham, Switzerland: Springer.

36. Article 29 Working Party. 2018 Guidelines on transparency under Regulation 2016/679 (WP260). See https://iapp.org/media/pdf/resource_center/wp29-transparency-12-12-17.pdf.

37. Mastelic T, Oleksiak A, Claussen H, Brandic I, Pierson JM, Vasilakos AV. 2015 Cloud computing: survey on energy efficiency. *ACM Comput. Surv.* **47**, 33. (doi:10.1145/2656204)

38. Cao Y, Yang J. 2015 Towards making systems forget with machine unlearning. In *Proc. of the 2015 IEEE Symp. on Security and Privacy (SP), San Jose, CA, 18–20 May 2015*, pp. 463–480. New York, NY: IEEE.

39. Barua D. 2016 A time to remember, a time to forget: user controlled, scalable, life long user modelling. PhD thesis, The University of Sydney, Sydney, Australia.

40. Google. 2017 Transparency report. See https://perma.cc/8DE4-AXBW.

41. Mahieu R, Asghari H, van Eeten M. 2017 Collectively exercising the right of access: individual effort, societal effect. In *Proc. of the GigaNet* (*Global Internet Governance Academic Network*) *Annual Symp, Geneva, Switzerland, 17 December 2017*. GigaNet.

42. Dwork C. 2008 Differential privacy: a survey of results. In *Proc. of the Int. Conf. on Theory and Applications of Models of Computation, Xi'an, China, 25–29 April 2008*, Dordrecht, The Netherlands: Springer.

43. Dwork C, McSherry F, Nissim K, Smith A. 2006 Calibrating noise to sensitivity in private data analysis. In *Proc. of the Theory of Cryptography: Third Theory of Cryptography Conf., TCC 2006, New York, NY, 4–7 March 2006* (eds S Halevi, T Rabin), pp. 265–284. Berlin, Germany: Springer.

44. Nissim K, Steinke T, Wood A, Altman M, Bembenek A, Bun M, Gaboardi M, O'Brien D, Vadhan S. In press. Differential privacy: a primer for a non-technical audience. *Vand. J. Ent. Technol. Law.*

45. Ji Z, Lipton ZC, Elkan C. 2014 Differential privacy and machine learning: a survey and review. (http://arxiv.org/abs/1412.7584)

46. Kasiviswanathan S, Lee H, Nissim K, Raskhodnikova S, Smith A. 2011 What can we learn privately? *SIAM J. Comput.* **40**, 793–826. (doi:10.1137/090756090)

47. Orlowski A. 2016 Apple pollutes data about you to protect your privacy. But it might not be enough. See https://www.theregister.co.uk/2016/06/21/apple_pollutes_your_data_differential_privacy/.

48. Doctorow C. 2013 Data protection in the EU: the certainty of uncertainty. *The Guardian*. See https://www.theguardian.com/technology/blog/2013/jun/05/data-protection-eu-anonymous.

49. Bradshaw T. 2016 Apple plays catch-up with imessage emojis. See https://www.ft.com/content/1d9a1aa6-31dc-11e6-bda0-04585c31b153 (accessed 11 October 2017).

50. Chaudhuri K, Monteleoni C, Sarwate AD. 2011 Differentially private empirical risk minimization. *J. Mach. Learn. Res.* **12**, 1069–1109.

51. Mironov I. 2012 On significance of the least significant bits for differential privacy. In *Proc. of the 2012 ACM Conf. on Computer and Communications Security, CCS '12, Raleigh, NC, 16–18 October 2012*, pp. 650–661. New York, NY: ACM.

52. Muralidhar K, Sarathy R. 2010 Does differential privacy protect terry gross' privacy? In *Privacy in Statistical Databases: UNESCO Chair in Data Privacy, International Conf., PSD 2010, Corfu, Greece, 22–24 September 2010* (eds J Domingo-Ferrer, E Magkos), pp. 200–209. Berlin, Germany: Springer.

53. Bambauer J, Muralidhar K, Sarathy R. 2013 Fool's gold: an illustrated critique of differential privacy. *Vand. J. Ent. Technol. Law* **16**, 701–755.

54. Yeom S, Fredrikson M, Jha S. 2017 The unintended consequences of overfitting: training data inference attacks. (http://arxiv.org/abs/1709.01604)

55. Gama J, Zliobaite I, Bifet A, Pechenizkiy M, Bouchachia A. 2013 A survey on concept drift adaptation. *ACM Comput. Surv.* **46**, 44. (doi:10.1145/2523813)